\documentclass[conference]{IEEEtran}

\IEEEoverridecommandlockouts

\usepackage{cite}
\usepackage{amsmath,amssymb,amsfonts}
\usepackage{algorithmic}
\usepackage{graphicx}
\usepackage{textcomp}
\usepackage{xcolor}
\usepackage{booktabs}
\usepackage{multirow}
\usepackage{tabularx, colortbl}
\usepackage{bm}
\usepackage{siunitx}
\usepackage{comment}
\usepackage{balance}

\usepackage[square,numbers]{natbib}

\usepackage{tikz}
\usetikzlibrary{external}
\usepackage{pgfplots}
\pgfplotsset{compat=1.9}
\usepgfplotslibrary{statistics}

\makeatletter
\let\MYcaption\@makecaption
\makeatother

\makeatletter
\let\@makecaption\MYcaption
\makeatother

\usepackage[font=footnotesize]{subcaption}
\usepackage[font=footnotesize]{caption}
\usepackage{balance}

\usepackage[letterpaper, top=54pt, bottom=54pt, left=54pt, right=54pt]{geometry}

\begin{document}

\title{\vspace*{18pt}\LARGE \bf On-The-Go Robot-to-Human Handovers with a Mobile Manipulator}

\author{Kerry He, Pradeepsundar Simini, Wesley Chan, Dana Kuli\'c, Elizabeth Croft, Akansel Cosgun \\Monash University, Australia}

\maketitle

\begin{abstract}

Existing approaches to direct robot-to-human handovers are typically implemented on fixed-base robot arms, or on mobile manipulators that come to a full stop before performing the handover. We propose ``on-the-go" handovers which permit a moving mobile manipulator to hand over an object to a human without stopping. The on-the-go handover motion is generated with a reactive controller that allows simultaneous control of the base and the arm. In a user study, human receivers subjectively assessed on-the-go handovers to be more efficient, predictable, natural, better timed and safer than handovers that implemented a ``stop-and-deliver" behavior. 
\end{abstract}


\section{Introduction} \label{sec:intro}

Today's robots are most commonly found in manufacturing industries and warehouse automation, and typically operate in isolation from humans. The introduction of robots that co-exist and interact with humans is creating an opportunity for new use cases, in both manufacturing and service robotics. One such use case is delivery robots. Robotic deliveries can either be performed indirectly by leaving the object in the vicinity of the human receiver \cite{choi2009hand,choi2022preemptive, quispe2017learning} or directly by handing over the object to them. The majority of research in robotic handovers is focused on direct handovers using fixed-base manipulators \cite{yang2021reactive,rosenberger2020object}. On the other hand, mobile robots allow for more control over when and how to approach a human receiver for fetch-and-carry tasks. While more recent works implement handovers on fixed-base robotic arms, earlier works feature humanoid robots \cite{chan2014determining, yamane2013synthesizing} and mobile manipulation platforms \cite{quispe2017learning,cakmak2011using, sisbot2010synthesizing,choi2009hand,chan2013human,vahrenkamp2016workspace,meyer2017hand}. In this work, we focus on robot-to-human handovers performed with a mobile manipulator.

Drawing parallels with human-to-human handovers, we would expect a human handing over an object to be able to perform the handover without completely stopping in their path. The handover is more efficient as the giver can move onto their next task quicker. This observation motivates us to transfer this behavior to robot givers. Our recent survey on robotic handovers revealed that none of the prior works proposed keeping the robot's base moving during the physical exchange stage of the handover \cite{ortenzi2021object}. We identified only two papers that considered the mobility of the handover agents. \citet{mainprice2012sharing} plans a path considering the mobility of a human receiver; however, the robot comes to a stop at the object transfer point. \citet{kupcsik2018learning} considers cases where the transfer point moves for a human receiver who may be walking or running; however, the handover is performed with a fixed-base robotic arm. In this paper, we consider mobile manipulators and exploit their full motion capability in robot-to-human handovers.
 
\begin{figure}[t]
\includegraphics[trim={160, 30, 50, 80},clip,width=0.99\columnwidth]{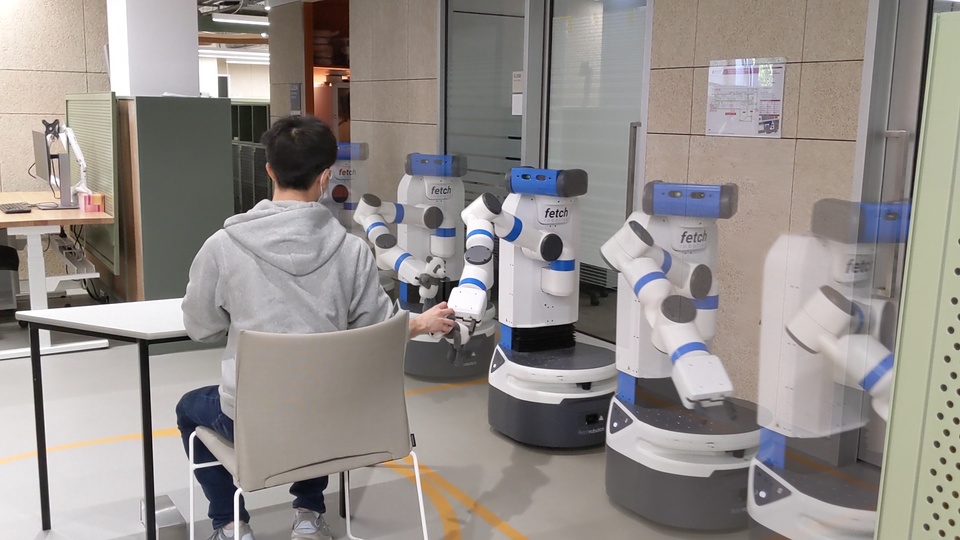}
\caption{An on-the-go handover. Human receivers assessed on-the-go handovers to be better than stop-and-deliver in most subjective metrics.}
\label{fig:intro_fig}
\vspace{-0.5cm}
\end{figure}

We present \textit{on-the-go} handovers, in which the robot's base keeps moving even during the physical exchange stage of the handover, as illustrated in Fig~\ref{fig:intro_fig}. To achieve on-the-go handovers, we adopt the reactive controller by \citet{haviland2021holistic}, which generates motions by moving the robot's base and its arm simultaneously. Our work investigates the subjective perceptions of human receivers for on-the-go handovers compared to stop-and-deliver handovers. To this end, we conduct an in-person user study in which we compare two handover behaviors (on-the-go vs stop-and-deliver) and two agent types (robot vs human).  The contributions of this work are two-fold:
\begin{itemize}
    \item We propose on-the-go handovers in which the robot's base continues to move during physical transfer. To the best of the authors' knowledge, this is the first reported study of mobile manipulation to perform object handovers throughout all phases of the handover.
    \item User studies that confirm the benefit of the on-the-go handovers compared to standard stop-and-deliver handovers in terms of subjective assessments by human receivers.
\end{itemize}



As on-the-go handovers are faster, we expect people to perceive it as more efficient and better timed. We also expect on-the-go handovers to be perceived as more natural and predictable as it conforms more closely to what a human would do, and more competent as it requires the robot to multi-task. In addition, we expect people to perceive the handover as safer as the motion is smooth and blended together. Furthermore, we expect that a robot giver will emphasize the differences between two handover behaviors more than a human giver would, as people are less familiar with robot handovers and will therefore be more attentive and critical towards it.

Therefore, we consider the following hypotheses:

\begin{itemize}
    \item \textbf{H1}: On-the-go handovers will be perceived more positively than stop-and-deliver handovers in all subjective measures.

    
    
    
    
    
    
    
    \item \textbf{H2}: Compared to stop-and-deliver handovers, on-the-go handovers will have a larger positive effect on subjective metrics for a robot giver compared to a human giver.
\end{itemize}

\section{Methodology} \label{sec:method}

\subsection{Robot Navigation and Control}

Robot-to-human handover experiments are implemented on a Fetch Mobile Manipulator which consists of a robotic arm mounted onto a non-holonomic mobile base, and a head which can pivot about two axes. 

A 2D occupancy map of the experimental environment is pre-computed by using a SLAM algorithm which fuses base laser scanner and depth camera measurements~\cite{zhu2019indoor}. This map is then used to perform on-line localization using a particle filter algorithm~\cite{zhu2019indoor}. The position of the seated human is assumed to be known within this map. 

The robot is controlled using an optimization-based reactive controller~\cite{haviland2021holistic}, which allows the simultaneous movement of the robot's base and the robotic arm. Target end-effector task-space velocities are computed to drive the end-effector in a straight line towards a waypoint defined relative to the mobile base. The optimal control problem is augmented with an additional control objective, formulated as a hard equality constraint, for the mobile base to meet a specified linear and angular velocity. These target velocities are calculated using a proportional controller to drive the base towards the final goal. The robot head is controlled independently using a similar proportional controller.

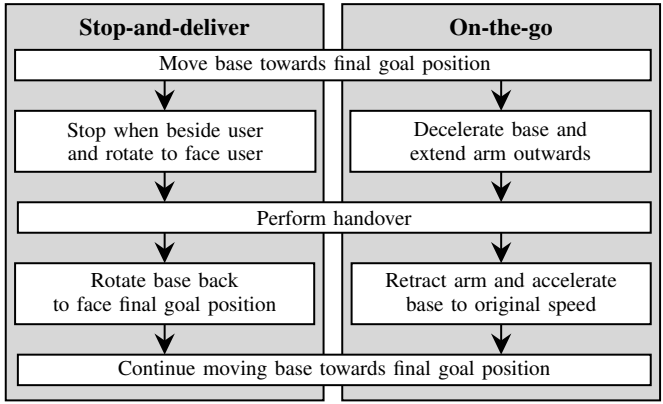
\begin{figure}[t]
\centering
\tikzset{every picture/.style={line width=0.75pt}} 

\begin{tikzpicture}[x=0.75pt,y=0.75pt,yscale=-1,xscale=1]

\draw  [fill={rgb, 255:red, 215; green, 215; blue, 215 }  ,fill opacity=1 ] (0,0) -- (160.29,0) -- (160.29,200) -- (0,200) -- cycle ;
\draw  [fill={rgb, 255:red, 215; green, 215; blue, 215 }  ,fill opacity=1 ] (169.71,0) -- (330,0) -- (330,200) -- (169.71,200) -- cycle ;
\draw  [fill={rgb, 255:red, 255; green, 255; blue, 255 }  ,fill opacity=1 ] (4.71,23.08) -- (325.29,23.08) -- (325.29,38.46) -- (4.71,38.46) -- cycle ;
\draw  [fill={rgb, 255:red, 255; green, 255; blue, 255 }  ,fill opacity=1 ] (4.71,53.85) -- (156.19,53.85) -- (156.19,84.62) -- (4.71,84.62) -- cycle ;
\draw  [fill={rgb, 255:red, 255; green, 255; blue, 255 }  ,fill opacity=1 ] (6.11,100) -- (325.29,100) -- (325.29,115.38) -- (6.11,115.38) -- cycle ;
\draw  [fill={rgb, 255:red, 255; green, 255; blue, 255 }  ,fill opacity=1 ] (4.71,130.77) -- (155.57,130.77) -- (155.57,161.54) -- (4.71,161.54) -- cycle ;
\draw  [fill={rgb, 255:red, 255; green, 255; blue, 255 }  ,fill opacity=1 ] (6.11,176.92) -- (325.29,176.92) -- (325.29,192.31) -- (6.11,192.31) -- cycle ;
\draw  [fill={rgb, 255:red, 255; green, 255; blue, 255 }  ,fill opacity=1 ] (173.81,53.85) -- (325.29,53.85) -- (325.29,84.62) -- (173.81,84.62) -- cycle ;
\draw    (80.14,38.46) -- (80.14,50.85) ;
\draw [shift={(80.14,53.85)}, rotate = 270] [fill={rgb, 255:red, 0; green, 0; blue, 0 }  ][line width=0.08]  [draw opacity=0] (10.72,-5.15) -- (0,0) -- (10.72,5.15) -- (7.12,0) -- cycle    ;
\draw  [fill={rgb, 255:red, 255; green, 255; blue, 255 }  ,fill opacity=1 ] (174.43,130.77) -- (325.29,130.77) -- (325.29,161.54) -- (174.43,161.54) -- cycle ;
\draw    (249.86,38.46) -- (249.86,50.85) ;
\draw [shift={(249.86,53.85)}, rotate = 270] [fill={rgb, 255:red, 0; green, 0; blue, 0 }  ][line width=0.08]  [draw opacity=0] (10.72,-5.15) -- (0,0) -- (10.72,5.15) -- (7.12,0) -- cycle    ;
\draw    (80.14,84.62) -- (80.14,97) ;
\draw [shift={(80.14,100)}, rotate = 270] [fill={rgb, 255:red, 0; green, 0; blue, 0 }  ][line width=0.08]  [draw opacity=0] (10.72,-5.15) -- (0,0) -- (10.72,5.15) -- (7.12,0) -- cycle    ;
\draw    (249.86,84.62) -- (249.86,97) ;
\draw [shift={(249.86,100)}, rotate = 270] [fill={rgb, 255:red, 0; green, 0; blue, 0 }  ][line width=0.08]  [draw opacity=0] (10.72,-5.15) -- (0,0) -- (10.72,5.15) -- (7.12,0) -- cycle    ;
\draw    (80.14,115.38) -- (80.14,127.77) ;
\draw [shift={(80.14,130.77)}, rotate = 270] [fill={rgb, 255:red, 0; green, 0; blue, 0 }  ][line width=0.08]  [draw opacity=0] (10.72,-5.15) -- (0,0) -- (10.72,5.15) -- (7.12,0) -- cycle    ;
\draw    (249.86,115.38) -- (249.86,127.77) ;
\draw [shift={(249.86,130.77)}, rotate = 270] [fill={rgb, 255:red, 0; green, 0; blue, 0 }  ][line width=0.08]  [draw opacity=0] (10.72,-5.15) -- (0,0) -- (10.72,5.15) -- (7.12,0) -- cycle    ;
\draw    (80.14,161.54) -- (80.14,173.92) ;
\draw [shift={(80.14,176.92)}, rotate = 270] [fill={rgb, 255:red, 0; green, 0; blue, 0 }  ][line width=0.08]  [draw opacity=0] (10.72,-5.15) -- (0,0) -- (10.72,5.15) -- (7.12,0) -- cycle    ;
\draw    (249.86,161.54) -- (249.86,173.92) ;
\draw [shift={(249.86,176.92)}, rotate = 270] [fill={rgb, 255:red, 0; green, 0; blue, 0 }  ][line width=0.08]  [draw opacity=0] (10.72,-5.15) -- (0,0) -- (10.72,5.15) -- (7.12,0) -- cycle    ;

\draw (160.96,30.77) node  [font=\footnotesize] [align=left] {Move base towards final goal position};
\draw (80.45,69.23) node  [font=\footnotesize] [align=left] {\begin{minipage}[lt]{85.74pt}\setlength\topsep{0pt}
\begin{center}
Stop when beside user\\and rotate to face user
\end{center}

\end{minipage}};
\draw (165.7,107.69) node  [font=\footnotesize] [align=left] {Perform handover};
\draw (80.14,146.15) node  [font=\footnotesize] [align=left] {\begin{minipage}[lt]{92.54pt}\setlength\topsep{0pt}
\begin{center}
Rotate base back\\to face final goal position
\end{center}

\end{minipage}};
\draw (165.7,184.62) node  [font=\footnotesize] [align=left] {Continue moving base towards final goal position};
\draw (249.55,69.23) node  [font=\footnotesize] [align=left] {\begin{minipage}[lt]{78.92pt}\setlength\topsep{0pt}
\begin{center}
Decelerate base and\\extend arm outwards
\end{center}

\end{minipage}};
\draw (249.86,146.15) node  [font=\footnotesize] [align=left] {\begin{minipage}[lt]{88.92pt}\setlength\topsep{0pt}
\begin{center}
Retract arm and accelerate\\base to original speed
\end{center}

\end{minipage}};
\draw (80.14,12.53) node  [font=\small] [align=left] {\begin{minipage}[lt]{108.99pt}\setlength\topsep{0pt}
\begin{center}
\textbf{Stop-and-deliver}
\end{center}

\end{minipage}};
\draw (249.86,12.53) node  [font=\small] [align=left] {\begin{minipage}[lt]{108.99pt}\setlength\topsep{0pt}
\begin{center}
\textbf{On-the-go}
\end{center}

\end{minipage}};

\end{tikzpicture}
\vspace{-0.4cm}
\caption{Flowchart diagram showing the behavior of the Fetch robot when performing a handover for both the stop-and-deliver and on-the-go strategies.}
\label{fig:seq}
\end{figure}

\subsection{Handover Behavior Implementation} 

To choose the robot arm configurations at the start and at the transfer point of the handover, we use the findings by \citet{cakmak2011using} who propose using contrast in the robot's actions for handovers. We adopt their idea of spatial contrast, by making the robot’s hand-over pose distinct from other things that the robot might do with an object in its hand. The tucked and handover configurations of the arm are fixed a priori, and defined relative to the robot base. These configurations are chosen to be within arm's length of the receiver, able to be quickly and smoothly transitioned to by the arm, and avoids colliding with the table and other obstacles.

\begin{figure}[t]
\centering
\input{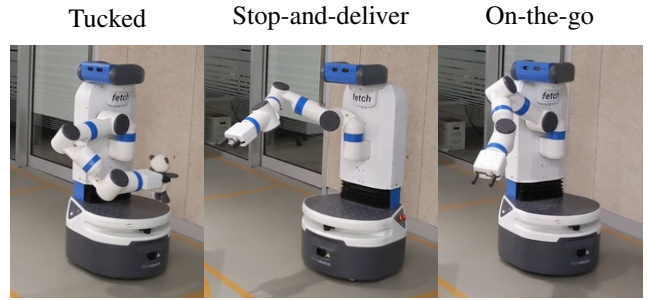}
\vspace{-0.5cm}
\caption{Different arm configurations the robot transitions between. The robot's head is pointed towards the user.}
\label{fig:arm}
\end{figure}

The robot begins with its arm in a tucked configuration (\SI{0.3}{\meter} forwards, \SI{0.6}{\meter} high) oriented away from the participant so that the object is not accessible by the participant  (as shown in Fig.~\ref{fig:arm}, left image). When a handover is performed, the end-effector is outstretched and tilted down slightly towards the participant (as shown in Fig.~\ref{fig:arm}, middle and right images). This change in configuration aims to maximize the spatial contrast between the robot's states to clearly communicate to the participant when the object is ready to be grasped. In addition, a simple gaze behavior is adopted where the robot looks at the participant while it approaches them and performs the handover. This was inspired by a user study by \citet{zheng2015impacts} that found that for robot-to-human handovers, people tend to reach for the offered object earlier when the robot provides a continual gaze cue towards the receiver's face. The robot then looks away towards the goal once the handover is complete.

The sequence of behaviors for the robot handovers is described below, and visualized in Fig.~\ref{fig:seq}:
\begin{itemize}
    \item \textbf{On-the-go (OtG)}: The robot begins with its arm in a tucked position and the mobile base begins moving down the corridor. Once it is within \SI{2.5}{\meter} of the participant, the mobile base gradually slows down to half its original speed, and the arm is extended towards the participant (\SI{0.3}{\meter} forwards, \SI{0.5}{\meter} to the side, \SI{0.8}{\meter} high). Once the robot has passed the participant, the mobile base speeds up to its original speed and retracts its arm to its original tucked configuration. 
    \item \textbf{Stop-and-deliver (S\&D)}: The robot begins with its arm in a tucked position and the mobile base begins moving down the corridor. Once it has arrived to the right of the participant, the base stops and rotates to face the participant. The arm is then outstretched to perform the handover (\SI{0.75}{\meter} forwards, \SI{0.25}{\meter} to the side, \SI{0.8}{\meter} high), then retracted when the handover is complete. The mobile base rotates to its original orientation, then continues moving towards its goal.
\end{itemize}

For human-to-human handovers, the human giver mimicked the robot's behavior as closely as possible. The role of the human giver was performed by the same experimenter in all experiments for consistency.

\subsection{Experimental Procedure}

\begin{figure}[t]
\centering
\tikzset{every picture/.style={line width=0.75pt}} 

\begin{tikzpicture}[x=0.75pt,y=0.75pt,yscale=-1,xscale=1]

\draw  [draw opacity=0][fill={rgb, 255:red, 243; green, 243; blue, 243 }  ,fill opacity=1 ] (105,50) -- (405,50) -- (405,100) -- (105,100) -- cycle ;
\draw  [draw opacity=0][fill={rgb, 255:red, 74; green, 74; blue, 74 }  ,fill opacity=1 ] (105,100) -- (196.3,100) -- (196.3,102.5) -- (105,102.5) -- cycle ;
\draw  [draw opacity=0][fill={rgb, 255:red, 74; green, 74; blue, 74 }  ,fill opacity=1 ] (313.7,100) -- (405,100) -- (405,102.5) -- (313.7,102.5) -- cycle ;
\draw  [fill={rgb, 255:red, 128; green, 128; blue, 128 }  ,fill opacity=1 ] (260,125) .. controls (260,119.48) and (264.48,115) .. (270,115) .. controls (275.52,115) and (280,119.48) .. (280,125) .. controls (280,130.52) and (275.52,135) .. (270,135) .. controls (264.48,135) and (260,130.52) .. (260,125) -- cycle ;
\draw  [fill={rgb, 255:red, 74; green, 144; blue, 226 }  ,fill opacity=1 ] (117.16,75) .. controls (117.16,70.67) and (120.67,67.16) .. (125,67.16) .. controls (129.33,67.16) and (132.84,70.67) .. (132.84,75) .. controls (132.84,79.33) and (129.33,82.84) .. (125,82.84) .. controls (120.67,82.84) and (117.16,79.33) .. (117.16,75) -- cycle ;
\draw    (132.84,75) -- (192,75) ;
\draw [shift={(195,75)}, rotate = 180] [fill={rgb, 255:red, 0; green, 0; blue, 0 }  ][line width=0.08]  [draw opacity=0] (10.72,-5.15) -- (0,0) -- (10.72,5.15) -- (7.12,0) -- cycle    ;
\draw  [fill={rgb, 255:red, 139; green, 87; blue, 42 }  ,fill opacity=1 ] (230,110) -- (250,110) -- (250,140) -- (230,140) -- cycle ;
\draw  [draw opacity=0][fill={rgb, 255:red, 74; green, 74; blue, 74 }  ,fill opacity=1 ] (105,47.5) -- (405,47.5) -- (405,50) -- (105,50) -- cycle ;
\draw  [color={rgb, 255:red, 0; green, 0; blue, 0 }  ,draw opacity=1 ][fill={rgb, 255:red, 0; green, 0; blue, 0 }  ,fill opacity=1 ] (262,122) .. controls (262,121.45) and (262.45,121) .. (263,121) .. controls (263.55,121) and (264,121.45) .. (264,122) .. controls (264,122.55) and (263.55,123) .. (263,123) .. controls (262.45,123) and (262,122.55) .. (262,122) -- cycle ;
\draw  [color={rgb, 255:red, 0; green, 0; blue, 0 }  ,draw opacity=1 ][fill={rgb, 255:red, 0; green, 0; blue, 0 }  ,fill opacity=1 ] (262,128) .. controls (262,127.45) and (262.45,127) .. (263,127) .. controls (263.55,127) and (264,127.45) .. (264,128) .. controls (264,128.55) and (263.55,129) .. (263,129) .. controls (262.45,129) and (262,128.55) .. (262,128) -- cycle ;
\draw   (105,40) -- (405,40) -- (405,160) -- (105,160) -- cycle ;
\draw  [color={rgb, 255:red, 255; green, 0; blue, 0 }  ,draw opacity=1 ][line width=3.75]  (379.61,69.39) -- (390.82,80.61)(390.82,69.39) -- (379.61,80.61) ;

\draw (125,58.5) node  [font=\footnotesize] [align=left] {\begin{minipage}[lt]{27.2pt}\setlength\topsep{0pt}
\begin{center}
Giver
\end{center}

\end{minipage}};
\draw (300,123.5) node  [font=\footnotesize] [align=left] {\begin{minipage}[lt]{27.2pt}\setlength\topsep{0pt}
\begin{center}
User
\end{center}

\end{minipage}};
\draw (240,148.5) node  [font=\footnotesize] [align=left] {\begin{minipage}[lt]{27.2pt}\setlength\topsep{0pt}
\begin{center}
Table
\end{center}

\end{minipage}};
\draw (255,91.5) node  [font=\footnotesize] [align=left] {\begin{minipage}[lt]{61.2pt}\setlength\topsep{0pt}
\begin{center}
Corridor
\end{center}

\end{minipage}};
\draw (385,58.5) node  [font=\footnotesize] [align=left] {\begin{minipage}[lt]{27.2pt}\setlength\topsep{0pt}
\begin{center}
Goal
\end{center}

\end{minipage}};

\end{tikzpicture}
\caption{Experimental setup of the user study.}
\label{fig:exp-setup}
\end{figure}
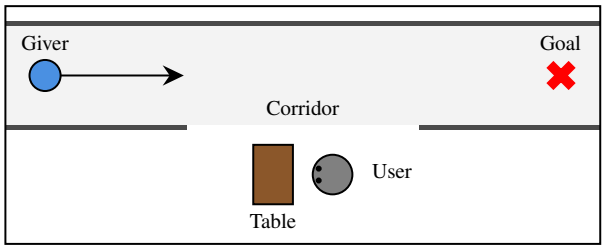

The participant, who was seated at a table next to the corridor as depicted in Fig.~\ref{fig:exp-setup}, was tasked to receive an object from the robot. The robot approaches the participant from their right side, following \citet{dautenhahn2006may} who found that the right side was the approach direction people are most comfortable with, with no significant differences in preferences being found between handedness. The giver starts at one end of the corridor where the participant cannot see them, so that they cannot initially know whether it is a robot or a human performing the handover. The giver is tasked with navigating to a goal at the opposite end of the corridor while performing the handover. A soft toy is used as the handover object and is loosely held in the gripper such that participants can easily take the object from the robot's gripper anytime, without the robot needing to detect and respond to when the person is taking the object.

\begin{figure}[t]
\centering
\input{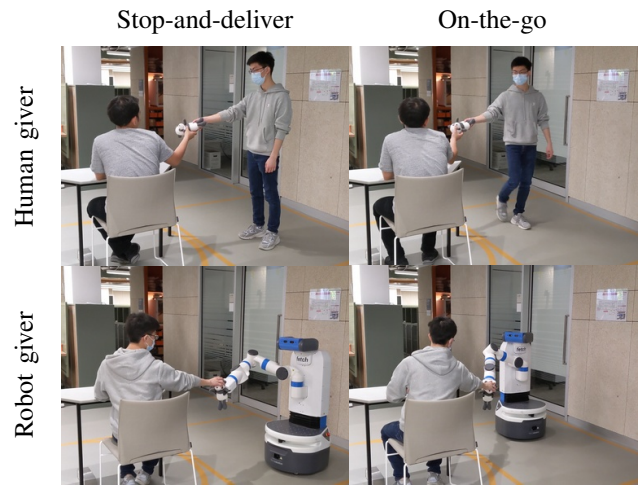}
\caption{Different handover scenarios that were studied, including human-to-human handovers (top), robot-to-human handovers (bottom), static handovers (left), and on-the-go handovers (right).}
\label{fig:2by2}
\end{figure}

Prior to the trials, the participant was told that there will be two handover behaviors and two handover givers, and that the handover agent will move down the corridor while performing the handover. Each participant is exposed to all $4$ types of handovers, as visualized in Fig.~\ref{fig:2by2}. The order in which participants experience these handovers is randomized, and they are not aware of which handover will happen before each trial. Each experimental trial began after the participant gave a verbal cue to the operators when they are ready. After each trial, the participant was asked to fill out a questionnaire regarding their experience with the latest trial. More details about the survey are given in the next section.

\subsection{Survey} \label{sec:survey}

To gauge the experiences of each participant during the experiments, participants were asked to fill out a survey after each trial to test the hypotheses listed in Sec.~\ref{sec:intro}. After each one of the 4 trials, the participants responded to the following 6 statements using a 5-point Likert scale:
\begin{enumerate}
    \item The giver was \textbf{efficient} in completing the handovers.
    
    \item The interaction with the giver felt \textbf{natural}.

    \item The giver's \textbf{timing} was appropriate.
    
    \item The giver was \textbf{competent} in giving objects to me.    
    
    \item I felt \textbf{safe} during the interaction.    

    \item I was able to \textbf{predict} what the giver was going to do.
\end{enumerate}
In addition, at the end of the $4$ trials, participants were asked about which handover behavior they preferred (On-The-Go/Stop-and-Deliver/No Preference):
\begin{itemize}
    \item Which \textbf{robot giver} handover behavior did you prefer? 
    \item Which \textbf{human giver} handover behavior did you prefer?    
\end{itemize}
At the end of the study, participants were asked to provide any additional comments they might have about their experience.

\section{Results} \label{sec:results}

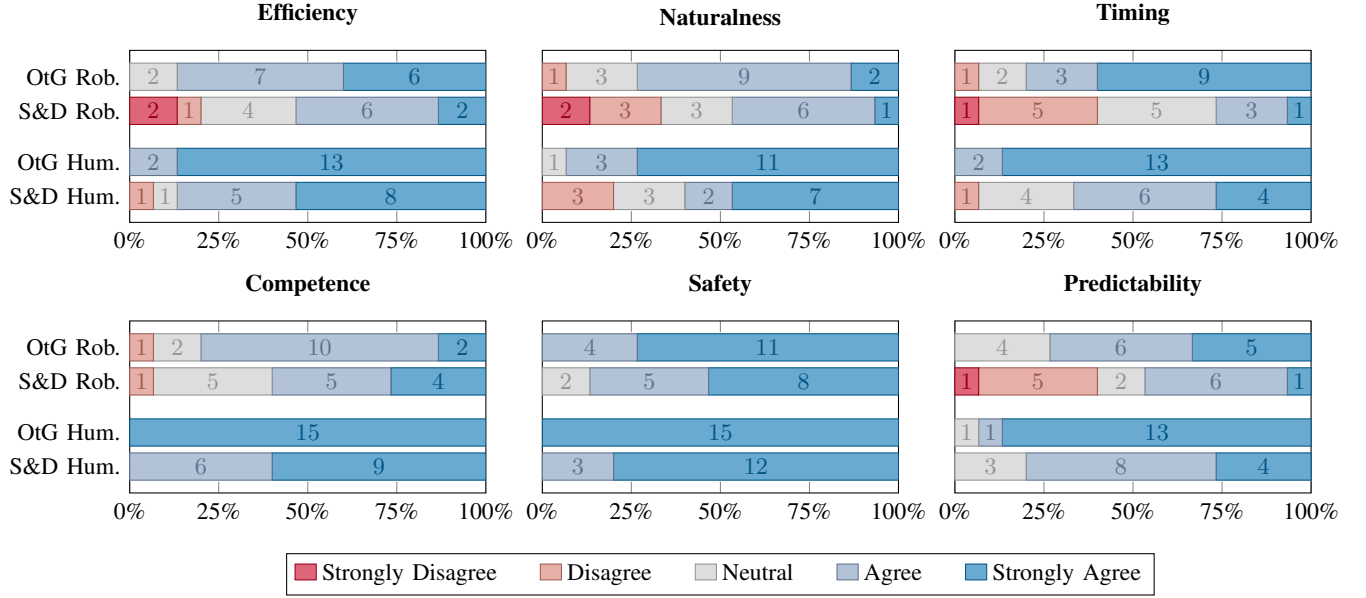
\begin{figure*}[h]
\centering
\begin{tikzpicture}
    \definecolor{r}{RGB}{202,0,32}
    \definecolor{rw}{RGB}{213, 123, 111}
    \definecolor{w}{RGB}{198, 198, 198}
    \definecolor{bw}{RGB}{127, 154, 188}
    \definecolor{b}{RGB}{5,113,176}

    \pgfplotsset{
       /pgfplots/bar  cycle  list/.style={/pgfplots/cycle  list={%
            {r!75!black,fill=r!60!white,mark=none},%
            {rw!75!black,fill=rw!60!white,mark=none},%
            {w!75!black,fill=w!60!white,mark=none},%
            {bw!75!black,fill=bw!60!white,mark=none},%
            {b!75!black,fill=b!60!white,mark=none},%
         }
       },
    }
    
    \tikzstyle{every node}=[font=\small]

    
        
        
        
        
        
    

    \begin{axis}[
        name=mainplot,
        xbar stacked,
        title=\textbf{Efficiency},
        nodes near coords,
        bar width=0.8,
        width = 0.355\textwidth,
        height = 0.2175\textwidth,
        xmin = 0, xmax = 15,
        enlarge y limits={abs=10pt},
        ytick={0,1,2.5,3.5},
        yticklabels={S\&D Hum., OtG Hum., S\&D Rob., OtG Rob.},  
        xtick={0,3.75,7.5,11.25,15},
        xticklabels={0\%,25\%,50\%,75\%,100\%},         
        legend style={at={(0.5,-0.20)}, anchor=north, legend columns=-1, /tikz/every even column/.append style={column sep=0.5cm}},
    ]
    
        \addplot coordinates
        {(0,3.5) (2,2.5) (0,1) (0,0)};
        
        \addplot coordinates
        {(0,3.5) (1,2.5) (0,1) (1,0)};
        
        \addplot coordinates
        {(2,3.5) (4,2.5) (0,1) (1,0)};
        
        \addplot coordinates
        {(7,3.5) (6,2.5) (2,1) (5,0)};
        
        \addplot coordinates
        {(6,3.5) (2,2.5) (13,1) (8,0)};      
        
    \end{axis}
    \begin{axis}[
        name=secondplot,
        title=\textbf{Naturalness},
        at={(mainplot.north east)},
        xshift=0.75cm,
        anchor=north west,    
        xbar stacked,
        nodes near coords,
        bar width=0.8,
        width = 0.355\textwidth,
        height = 0.2175\textwidth,
        xmin = 0, xmax = 15,
        enlarge y limits={abs=10pt},
        ytick={0,1,2.5,3.5},
        yticklabels={,,,},  
        xtick={0,3.75,7.5,11.25,15},
        xticklabels={0\%,25\%,50\%,75\%,100\%},         
        legend style={at={(0.5,-0.20)}, anchor=north, legend columns=-1, /tikz/every even column/.append style={column sep=0.5cm}},
    ]
    
        \addplot coordinates
        {(0,3.5) (2,2.5) (0,1) (0,0)};
        
        \addplot coordinates
        {(1,3.5) (3,2.5) (0,1) (3,0)};
        
        \addplot coordinates
        {(3,3.5) (3,2.5) (1,1) (3,0)};
        
        \addplot coordinates
        {(9,3.5) (6,2.5) (3,1) (2,0)};
        
        \addplot coordinates
        {(2,3.5) (1,2.5) (11,1) (7,0)};    
        
    \end{axis}
    \begin{axis}[
        name=thirdplot,
        title=\textbf{Timing},
        at={(secondplot.north east)},
        xshift=0.75cm,
        anchor=north west,      
        xbar stacked,
        nodes near coords,
        bar width=0.8,
        width = 0.355\textwidth,
        height = 0.2175\textwidth,
        xmin = 0, xmax = 15,
        enlarge y limits={abs=10pt},
        ytick={0,1,2.5,3.5},
        yticklabels={,,,},  
        xtick={0,3.75,7.5,11.25,15},
        xticklabels={0\%,25\%,50\%,75\%,100\%},         
        legend style={at={(0.5,-0.20)}, anchor=north, legend columns=-1, /tikz/every even column/.append style={column sep=0.5cm}},
    ]
    
        \addplot coordinates
        {(0,3.5) (1,2.5) (0,1) (0,0)};
        
        \addplot coordinates
        {(1,3.5) (5,2.5) (0,1) (1,0)};
        
        \addplot coordinates
        {(2,3.5) (5,2.5) (0,1) (4,0)};
        
        \addplot coordinates
        {(3,3.5) (3,2.5) (2,1) (6,0)};
        
        \addplot coordinates
        {(9,3.5) (1,2.5) (13,1) (4,0)};    
        
    \end{axis}
    \begin{axis}[
        at={(mainplot.below south west)},
        title=\textbf{Competence},
        yshift=-0.85cm,
        anchor=north west,
        xbar stacked,
        nodes near coords,
        bar width=0.8,
        width = 0.355\textwidth,
        height = 0.2175\textwidth,
        xmin = 0, xmax = 15,
        enlarge y limits={abs=10pt},
        ytick={0,1,2.5,3.5},
        yticklabels={S\&D Hum., OtG Hum., S\&D Rob., OtG Rob.},  
        xtick={0,3.75,7.5,11.25,15},
        xticklabels={0\%,25\%,50\%,75\%,100\%},         
        legend style={at={(0.5,-0.20)}, anchor=north, legend columns=-1, /tikz/every even column/.append style={column sep=0.5cm}},
    ]
    
        \addplot coordinates
        {(0,3.5) (0,2.5) (0,1) (0,0)};
        
        \addplot coordinates
        {(1,3.5) (1,2.5) (0,1) (0,0)};
        
        \addplot coordinates
        {(2,3.5) (5,2.5) (0,1) (0,0)};
        
        \addplot coordinates
        {(10,3.5) (5,2.5) (0,1) (6,0)};
        
        \addplot coordinates
        {(2,3.5) (4,2.5) (15,1) (9,0)};     
        
    \end{axis}
    \begin{axis}[
        title=\textbf{Safety},
        at={(secondplot.below south west)},
        yshift=-0.85cm,
        anchor=north west,    
        xbar stacked,
        nodes near coords,
        bar width=0.8,
        width = 0.355\textwidth,
        height = 0.2175\textwidth,
        xmin = 0, xmax = 15,
        enlarge y limits={abs=10pt},
        ytick={0,1,2.5,3.5},
        yticklabels={,,,},   
        xtick={0,3.75,7.5,11.25,15},
        xticklabels={0\%,25\%,50\%,75\%,100\%},         
        legend style={at={(0.5,-0.35)}, anchor=north, legend columns=-1, /tikz/every even column/.append style={column sep=0.5cm}},
    ]
    
        \addplot coordinates
        {(0,3.5) (0,2.5) (0,1) (0,0)};
        
        \addplot coordinates
        {(0,3.5) (0,2.5) (0,1) (0,0)};
        
        \addplot coordinates
        {(0,3.5) (2,2.5) (0,1) (0,0)};
        
        \addplot coordinates
        {(4,3.5) (5,2.5) (0,1) (3,0)};
        
        \addplot coordinates
        {(11,3.5) (8,2.5) (15,1) (12,0)};       
        
        \legend{Strongly Disagree, Disagree, Neutral, Agree, Strongly Agree}
        
    \end{axis}
    \begin{axis}[
        title=\textbf{Predictability},
        at={(thirdplot.below south west)},
        yshift=-0.85cm,
        anchor=north west,   
        xbar stacked,
        nodes near coords,
        bar width=0.8,
        width = 0.355\textwidth,
        height = 0.2175\textwidth,
        xmin = 0, xmax = 15,
        enlarge y limits={abs=10pt},
        ytick={0,1,2.5,3.5},
        yticklabels={,,,},  
        xtick={0,3.75,7.5,11.25,15},
        xticklabels={0\%,25\%,50\%,75\%,100\%},         
        legend style={at={(0.5,-0.20)}, anchor=north, legend columns=-1, /tikz/every even column/.append style={column sep=0.5cm}},
    ]
    
        \addplot coordinates
        {(0,3.5) (1,2.5) (0,1) (0,0)};
        
        \addplot coordinates
        {(0,3.5) (5,2.5) (0,1) (0,0)};
        
        \addplot coordinates
        {(4,3.5) (2,2.5) (1,1) (3,0)};
        
        \addplot coordinates
        {(6,3.5) (6,2.5) (1,1) (8,0)};
        
        \addplot coordinates
        {(5,3.5) (1,2.5) (13,1) (4,0)};      
        
    \end{axis}

\end{tikzpicture}
\vspace{-0.4cm}
\caption{Summary of the raw data obtained from the survey filled out by $15$ participants, comparing on-the-go (OtG) and stop-and-deliver (S\&D) handovers. The exact number of responses per Likert option is shown in each respective bar.}
\label{fig:barplot}
\vspace{-0.4cm}
\end{figure*}

We recruited $15$ participants from the lab and University premises\footnote[1]{No external participants could be recruited due to the university COVID-19 policy. This study has been approved by the Monash University Human Research Ethics Committee (Application ID: 31765).}, including $11$ male and $4$ female participants between the ages $20$ and $31$ ($\mu=22.9$, $\sigma=3.56$). All participants had some prior experience working with robotic platforms. There were $3$ out of the $60$ trials that were repeated for the following reasons: an error in the experimental setup, a localization error of the robot, and a handover failure due to a participant's uncertainty about how to interact with the robot on their first trial (a stop-and-deliver robot-to-human handover). All following statistical tests are performed to a $5\%$ significance level. The distribution of all responses are shown in Fig.~\ref{fig:barplot} and the user preferences are shown in Fig.~\ref{fig:pref}.

\subsection{Analysis}

Due to the ordinal nature of the Likert scale variables, a non-parametric test was applied. To test \textbf{H1}, a single-tailed Pratt modified Wilcoxon signed-rank test was ran for paired differences between perceptions of the on-the-go and stop-and-deliver handovers. As we believe that a robot agent compared to a human agent fundamentally changes the nature of the handover, we do not treat our experiment as a $2\times2$ factorial design, and instead analyze the robot-to-human and human-to-human experiment results independently. The statistical test results are summarised in Table~\ref{table:h1h6}.

\begin{table}[h]
\small
\begin{tabular*}{\columnwidth}{@{}l@{\extracolsep{\fill}}ccccc@{}}
\toprule
 & \multirow{2}{*}{$\bm{H_a}$} & \multicolumn{2}{c}{\textbf{Robot}} & \multicolumn{2}{c}{\textbf{Human}} \\ \cmidrule(l){3-6}
\multicolumn{1}{c}{} & & \textbf{W(15)} & $\bm{p}$ & \textbf{W(15)} & $\bm{p}$ \\ \midrule
\textbf{Efficiency} & OtG\textgreater{}S\&D & 87.0 & \textbf{0.021} & 81.0 & \textbf{0.016} \\
\textbf{Naturalness} & OtG\textgreater{}S\&D & 80.0 & \textbf{0.020} & 85.0 & \textbf{0.041} \\
\textbf{Timing} & OtG\textgreater{}S\&D & 114.0 & \textbf{0.001} & 106.5 & \textbf{0.002} \\
\textbf{Competence} & OtG\textgreater{}S\&D & 55.0 & 0.291 & 75.0 & \textbf{0.007} \\
\textbf{Safety} & OtG\textgreater{}S\&D & 63.0 & \textbf{0.049} & 42.0 & \textbf{0.042} \\
\textbf{Predictability} & OtG\textgreater{}S\&D & 90.5 & \textbf{0.008} & 101.0 & \textbf{0.003} \\ \bottomrule
\end{tabular*}
\caption{Wilcoxon signed-rank test results for robot handover behavior hypotheses H1. Significant results are indicated in \textbf{bold}.}
\label{table:h1h6}
\end{table}

For human-to-human handovers, on-the-go handovers were rated significantly better than stop-and-deliver handovers with respect to all survey questions. This was expected, as it is natural for people keep moving as they are handing over objects.

For robot-to-human handovers, on-the-go handovers were found to be significantly more \textbf{efficient}, \textbf{natural}, \textbf{safer}, \textbf{predictable}, and to have better \textbf{timing}. These results affirm \textbf{H1} to an extent. Interestingly, we did not find a statistically significant improvement in the perceived \textbf{competence} of the robot, therefore \textbf{H1} was not fully affirmed.

To test \textbf{H2}, we define two new sets of variables as the difference between on-the-go and stop-and-deliver handover ratings for each of the robot and human agent experiments. Single-tailed Pratt modified Wilcoxon signed-rank tests are then used to test for \textbf{H2} for each subjective metric. The results for this statistical test are summarized in Table~\ref{table:h8}. No subjective metric featured a statistically significant larger positive effect when using on-the-go handovers over stop-and-deliver handovers when compared between the two handover agents. 

\begin{table}[h]
\small
\begin{tabular*}{\columnwidth}{@{}l@{\extracolsep{\fill}}ccc@{}}
\toprule
\multicolumn{1}{c}{} & $\bm{H_a}$ & \textbf{W(15)} & \textbf{$\bm{p}$} \\ \midrule
\textbf{Efficiency} & R(OtG$-$S\&D)\textgreater{}H(OtG$-$S\&D) & 71.5 & 0.196 \\
\textbf{Naturalness} & R(OtG$-$S\&D)\textgreater{}H(OtG$-$S\&D) & 55.5 & 0.488 \\
\textbf{Timing} & R(OtG$-$S\&D)\textgreater{}H(OtG$-$S\&D) & 75.0 & 0.094 \\
\textbf{Competence} & R(OtG$-$S\&D)\textgreater{}H(OtG$-$S\&D) & 38.0 & 0.844 \\
\textbf{Safety} & R(OtG$-$S\&D)\textgreater{}H(OtG$-$S\&D) & 40.0 & 0.303 \\
\textbf{Predictability} & R(OtG$-$S\&D)\textgreater{}H(OtG$-$S\&D) & 69.0 & 0.270 \\ \bottomrule
\end{tabular*}
\caption{Wilcoxon signed-rank test results for H2. No result was statistically significant, which means that handovers with robotic givers did not benefit more from the on-the-go feature compared to human givers.}
\label{table:h8}
\end{table}

\subsection{Discussion}

From the survey data shown in Fig.~\ref{fig:barplot}, we observe that all participants felt safe with the on-the-go robot-to-human handover (all either agree or strongly agree). We believe factors including the robot slowing down when it is closer to the participant, and approaching from the side as opposed to a front-on approach, helped to achieve this result. However, out of all of the subjective metrics for on-the-go robot-to-human handovers, participants rated the handover's naturalness and competence the lowest, with only two participants strongly agreeing that the robot exhibited each of these traits. This is likely influenced by the specific handover behavior implementation used, such as the timing of the handover, positioning of the handover point, gaze patterns, and other possible expected verbal or non-verbal cues. Future work is recommended to fine-tune the behavior of the on-the-go handover make the handover feel more natural and competent.

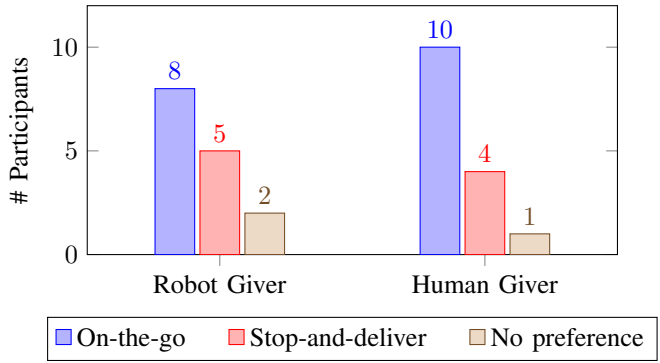
\begin{figure}[t]
\centering
\begin{tikzpicture}
    \begin{axis}[
        ybar,
        enlarge x limits=0.5,
        legend style={at={(0.5,-0.25)},
        anchor=north,legend columns=-1,
        /tikz/every even column/.append style={column sep=0.5cm}},
        ylabel={\# Participants},
        symbolic x coords={Robot Giver,Human Giver},
        xtick=data,
        nodes near coords,
        nodes near coords align={vertical},
        bar width=15pt,
        xtick pos=left,
        width = 0.485\textwidth,
        height = 0.275\textwidth,
        ymin = 0, ymax = 12,
        legend image code/.code={\draw [#1] (0cm,-0.075cm) rectangle (0.2cm,0.125cm); },
    ]
    
    \addplot coordinates {(Robot Giver,8) (Human Giver,10)};
    
    \addplot coordinates {(Robot Giver,5) (Human Giver,4)};
    
    \addplot coordinates {(Robot Giver,2) (Human Giver,1)};
    
    \legend{On-the-go,Stop-and-deliver,No preference}
    
    \end{axis}
\end{tikzpicture}
\vspace{-0.4cm}
\caption{Distribution of handover preferences between $15$ participants.}
\label{fig:pref}
\end{figure}

We also observe that receivers subjectively preferred human givers over robot givers ($p\leq0.023$ for all subjective measures and givers using a Pratt modified Wilcoxon signed-rank test). This is somewhat in contrast to the study by \citet{unhelkar2014comparative} which did not find a significant difference in the perceptions of fluency between human and robotic givers. As opposed to their study, we did not ask about the fluency of the interaction in our survey, and their platform was a mobile robot without an arm which might change the perception of the receivers.

Our hypothesis tests show that robot handover competence was the only metric that did not improve when comparing on-the-go handovers to stop-and-deliver handovers. This may be due to participants perceiving the robot givers as equally less competent compared to human givers, due to drawbacks shared between both robot handover behaviors such as not reactively adjusting to the receiver's preferences. However, further work should be conducted to verify this hypothesis. All other subjective metrics favoured on-the-go handovers. 


Although the results regarding the subjective metrics suggest that on-the-go handovers are superior overall, this preference is less clear when participants were asked explicitly which handover they preferred, as shown in Fig.~\ref{fig:pref}. Only a slight majority of people preferred robot on-the-go handovers (OtG: $54\%$, S\&D: $33\%$, No pref.: $13\%$). A possible explanation for this is that the preferred handover is also dependant on the specific task, as posited by \citet{martinson2017towards}. This would be a factor participants consider for their overall handover preference, but not when responding to Likert items regarding the specific task used in the user study. We received the following comments which support this:
\begin{itemize}
    \item ``\textit{[It] felt like each handover is good for a different purpose. [I] preferred [the on-the-go] handover for this particular [object].}"

    \item ``\textit{If the robot was bringing me a plate of food, I would have preferred the [stop-and-deliver handover] since its safer.}"
\end{itemize}
Therefore, the results regarding subjective metrics may only be relevant to this specific handover task. Future work generalizing on-the-go handovers to more tasks may consider the role of object semantics for handovers.

Another observation is that participants were influenced by differences in handover behaviors not directly arising from the on-the-go-stop-and-deliver handover dichotomy. Three participants commented that the position the agents stopped at to perform the stop-and-deliver handover (directly to the right of the participant) felt awkward, and would have preferred the robot to stop slightly earlier. However, this may also be an advantage of on-the-go robots, as participants can choose where to perform the handover along a continuum. Three participants commented that they would have preferred the robot on-the-go handover to reach its arm a bit further or higher. Future work may be required to remove these slight unwanted behavioral differences to ensure there are no confounding effects in the experiment. Alternatively, a more intelligent method to adaptively choose the handover location based on the user's position and/or preference could be beneficial to people's perception of the handover.



\textbf{H2} was not supported by the results, showing that there was not a significant difference between people's perception of the improvement of on-the-go handovers over stop-and-deliver handovers between human and robot agents. As opposed to our original justification for \textbf{H2}, this may instead suggest that the benefits of a on-the-go handover are as relevant for a human agent as they are for a robot agent.


\section{Conclusions, Limitations and Future Work} \label{sec:conclusion}

First stopping, and then delivering the object is the standard way performing handovers with mobile manipulation platforms in the existing literature. On the other hand, ``on-the-go" handovers, where the robot does not come to a stop during physical handover, can be more efficient since the giver can move to its next task earlier. In this paper, we conduct a user study to investigate the subjective perception of on-the-go handovers compared to stop-and-deliver handovers, from the receivers' perspective.

Our user study results showed that receivers subjectively assessed robot-to-human on-the-go handovers as more \textbf{efficient}, \textbf{natural}, \textbf{safer} and \textbf{predictable} and to have a better \textbf{timing} compared to stop-and-deliver handovers. This suggests that robot-to-human handovers should be performed as on-the-go for scenarios similar to our experiment, however, further qualitative research is needed to understand when and where on-the-go handovers are suitable \cite{quispe2017learning}.
    

A limitation of this work is the assumption that the human receiver always pays attention to the handover task. In real-world deployments, users would divide their attention between the handover and other activities they are doing, which might be as simple as checking their phones~\cite{martinson2017towards}. Catching the attention of the handover target is more important for on-the-go handovers compared to fixed manipulators, because if the robot cannot catch the attention of the human receiver, then the on-the-go handover would be reduced to a stop-and-deliver behavior. Detecting if the user is ready for the handover would be important for real-world situations~\cite{kwan2020gesture}. The robot can also actively seek to increase the chance of engagement by using gaze cues~\cite{moon2014meet}, better approach angles~\cite{macharet2013learning, satake2009approach}, predictable arm motions~\cite{dragan2013legibility, chan2021experimental}, audio cues~\cite{yilmazyildiz2016review}, or a combination thereof~\cite{shi2013model}. If waiting for the user to complete the handover is deemed to be detrimental to the efficiency of a robot, then indirect handovers can be a way to circumvent this issue \cite{choi2022preemptive}, where the robot places the object on a table~\cite{newbury2021learning}.

Other limitations of our work include the assumption that the receiver's position is known, and that we don't detect if the user actually picked up the object from the robot's gripper. These limitations can be addressed with perception modules for detecting people \cite{kwan2020gesture}, predicting the object transfer point~\cite{nemlekar2019object}, and checking grip forces to detect if the object has been picked up~\cite{chan2012grip}.

Future work should consider situations where the human receiver is walking instead of sitting, or extend our previous work on human-to-robot handovers~\cite{rosenberger2020object} into on-the-go handovers. For such handovers, the robot would need to keep the target object in-sight as much as possible~\cite{he2022visibility}, and a faster gripper might be needed. It may also be useful to utilize a projection method such as augmented reality to make the robot's intent accessible to the users~\cite{newbury2021visualizing}.

\section{Acknowledgement}
This project was supported by the Australian Research Council (ARC) Discovery Project Grant DP200102858.

\balance

\bibliographystyle{IEEEtranN}
\bibliography{refs}

\end{document}